\documentclass[10pt,twocolumn,letterpaper]{article}

\usepackage{wacv}              

\usepackage{graphicx}
\usepackage{amsmath}
\usepackage{amssymb}
\usepackage{booktabs}
\usepackage{array}
\usepackage{subcaption}
\usepackage{multirow}
\usepackage{arydshln}
\usepackage{tabulary}
\usepackage{colortbl}
\usepackage{xcolor}
\usepackage{enumitem}

\usepackage[pagebackref,breaklinks,colorlinks]{hyperref}

\usepackage[capitalize]{cleveref}
\crefname{section}{Sec.}{Secs.}
\Crefname{section}{Section}{Sections}
\Crefname{table}{Table}{Tables}
\crefname{table}{Tab.}{Tabs.}

\def\wacvPaperID{2487} 
\def\confName{WACV}
\def\confYear{2025}

\begin{document}
\raggedbottom
\title{Style-Pro: Style-Guided Prompt Learning for Generalizable Vision-Language Models}

\author{Niloufar Alipour Talemi \quad Hossein Kashiani \quad Fatemeh Afghah\\
Clemson University\\
{\tt\small  \{nalipou, hkashia, fafghah\}@clemson.edu}
}
\maketitle

\begin{abstract}
Pre-trained Vision-language (VL) models, such as CLIP, have shown significant generalization ability to downstream tasks, even with minimal fine-tuning. While prompt learning has emerged as an effective strategy to adapt pre-trained VL models for downstream tasks, current approaches frequently encounter severe overfitting to specific downstream data distributions. This overfitting constrains the original behavior of the VL models to generalize to new domains or unseen classes, posing a critical challenge in enhancing the adaptability and generalization of VL models. To address this limitation, we propose Style-Pro, a novel style-guided prompt learning framework that mitigates overfitting and preserves the zero-shot generalization capabilities of CLIP. Style-Pro employs learnable style bases to synthesize diverse distribution shifts, guided by two specialized loss functions that ensure style diversity and content integrity. Then, to minimize discrepancies between unseen domains and the source domain, Style-Pro maps the unseen styles into the known style representation space as a weighted combination of style bases. Moreover, to maintain consistency between the style-shifted prompted model and the original frozen CLIP, Style-Pro introduces consistency constraints to preserve alignment in the learned embeddings, minimizing deviation during adaptation to downstream tasks. Extensive experiments across 11 benchmark datasets demonstrate the effectiveness of Style-Pro, consistently surpassing state-of-the-art methods in various settings, including base-to-new generalization, cross-dataset transfer, and domain generalization.

\end{abstract}

\section{Introduction}
\label{sec:intro}
Vision-language (VL) models like CLIP \cite{radford2021learning} have exhibited remarkable generalization capabilities across various downstream tasks, including few-shot image recognition \cite{kim2021adapt, gao2024clip, bose2024stylip}, object detection \cite{feng2022promptdet}, and image segmentation \cite{ding2022decoupling, he2023clip}. These models are trained on millions of image-text pairs through contrastive loss, producing a well-aligned joint embedding space for vision and language. For instance, CLIP employs the contrastive learning strategy on a web-scale dataset with 400 million image-text
pairs that achieves impressive zero-shot
generalization ability by providing text prompts such as “a photo of a [class]”. However, due to their large size, these models can be difficult to fine-tune for smaller tasks, such as few-shot learning, without compromising their generalization capabilities.

\begin{figure}[t]

    \centering 
    \includegraphics[scale=1.4]{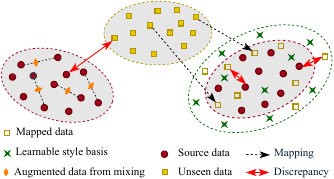}
    \caption{Illustration of the proposed style shift learning approach. Simply mixing feature statistics \cite{zhou2021domain, wang2022feature} from the source domain does not generate styles sufficiently distinct from the source domain. Style-Pro addresses this by employing a learnable set of style bases to synthesize style bases beyond the source domain. Furthermore, mapping unseen styles into the style representation space as a weighted combination of these style bases reduces the discrepancy between unseen domains and the source domain.}
    \label{fig:First}%
\end{figure}

One straightforward approach is prompt engineering, where, instead of fine-tuning the entire model, which may result in the loss of valuable knowledge gained during large-scale pre-training, the focus is on identifying effective hand-crafted templates for each task \cite{radford2021learning}. However, this technique demands considerable manual effort, as designing effective prompts requires deep expert knowledge and substantial time investment. To address this, recent works often incorporate learnable prompts into the text encoder \cite{zhou2022conditional, zhou2022learning, bulat2023lasp}, the image encoder \cite{rao2022denseclip, tsimpoukelli2021multimodal}, or both \cite{khattak2023maple}. In such approaches, the prompts are optimized during training, while the pre-trained VL models remain unchanged. Prompt learning has become increasingly popular, as it provides an effective way to customize pre-trained VL models for various downstream tasks. Despite advancements in few-shot fine-tuning, optimizing prompts for task-specific objectives often causes the model to overfit to the training samples \cite{zhou2022conditional, khattak2023maple}. This overfitting limits the model's generalization to new domains or even new classes within the same domain, which has remained a major challenge for
adapting the VL models \cite{zhou2022conditional}.

This work introduces a novel style-guided prompt learning framework called Style-Pro to address the challenges of domain bias in prompt tuning for CLIP. Style-Pro achieves this through two complementary strategies. First, to enhance model generalization and robustness to out-of-distribution (OOD) data, we introduce a style shift learning approach within the prompt-tuning framework. Unlike traditional image-based augmentation methods \cite{cubuk2020randaugment, hong2021stylemix}, which often fail to capture a realistic and comprehensive range of domain variations, our proposed approach leverages learnable style bases to extend the diversity of training distribution. In response to the challenges posed by a limited number of samples per class and the lack of samples from target domains, Style-Pro employs two synergistic phases in its style shift learning module. As illustrated in Fig. \ref{fig:First}, rather than simply mixing source feature statistics, Style-Pro facilitates the exploration of novel style bases beyond the source distribution during training without requiring additional images. This strategy effectively expands the source domain and minimizes discrepancies between the source and unseen domains. We introduce two new loss functions to synthesize style bases that incorporate various domains while preserving the content information of features. The first one encourages diversity among the style bases by ensuring they spread across a hyperspherical feature space. The second loss function provides the integrity of content information when combining style and content into a new feature. In addition to generating new style bases, Style-Pro maps unseen styles into the style representation space as a weighted combination of style bases (see Fig. \ref{fig:First}). The similarity distance between styles serves as the weighting factor, leading to improved adaptability to unseen domains.  

Second, to maintain consistency between the style-shifted prompted model and the original frozen CLIP model, Style-Pro enforces alignment by constraining both the text and image encoders. To be more specific, we transfer knowledge from the frozen encoders to the adaptable ones, preserving the robust generalization capabilities of the pre-trained VL model while effectively adapting to a new task within a few-shot learning context. To fully leverage the guidance from the frozen model, we apply two distinct constraints. The first constraint focuses on the difference between the feature representations of the frozen and prompted encoders, while the second is prediction-based, aiming to improve alignment by fostering synergy between the text and vision encoders. These constraints ensure that the embeddings of the prompted model remain closely aligned with the frozen model, even as it learns new tasks. In summary, the main contributions of this work include:
\begingroup
    \setlength{\leftmargini}{1.8pt}
    \setlength{\labelsep}{1.9 pt}
 \begin{itemize}[align=left,leftmargin=*]
\item {We present Style-Pro, a fine-tuning method for CLIP that mitigates overfitting by constraining the prompted model to remain aligned with the frozen CLIP. This approach facilitates efficient learning from limited data while preserving zero-shot generalization capabilities.}
\item We propose a novel style shift learning approach that synthesizes a diverse range of distinctive styles through learnable style bases. Style-Pro further improves adaptability to unseen domains by representing unseen styles as a weighted combination of these style bases. 
\item We conduct a comprehensive quantitative analysis on 11 popular image classification benchmarks, demonstrating the effectiveness and robustness of the Style-Pro framework across base-to-novel generalization, cross-dataset transfer, and domain generalization tasks. 
\end{itemize}

\endgroup

\section{Related Work}
\subsection{Vision-Language Models.}
VL models \cite{jia2021scaling, radford2021learning, yao2021filip} integrate visual and textual modalities to encode comprehensive multi-modal representations. These models are typically pre-trained on extensive datasets; for instance, CLIP \cite{radford2021learning} and ALIGN \cite{jia2021scaling} are trained on approximately 400 million and 1 billion image-text pairs, respectively. Through self-supervised learning, VL models develop joint image-language representations that significantly enhance their capacity for representation learning. As a result, these models exhibit outstanding performance across a wide range of tasks, including few-shot \cite{chen2022plot, lu2022prompt, naeem2022i2dformer} and zero-shot visual recognition \cite{radford2021learning}. Nonetheless, adapting these foundational models to specific downstream tasks without diminishing their inherent generalization capabilities remains a significant challenge. In this study, we introduce a novel fine-tuning approach for VL models that mitigates the issue of overfitting and enhances generalization by employing a style-guided prompt learning framework.
\subsection{Multi-Modal Prompt Tuning for VL Models}
Prompt learning has emerged as an efficient technique for fine-tuning large-scale models by integrating learnable embeddings, known as prompt tokens, into the model inputs \cite{lu2022prompt, derakhshani2023bayesian, zhou2022conditional}. Due to its parameter efficiency and fast convergence, prompt learning is a compelling method for adapting foundational models like CLIP \cite{radford2021learning} for both vision and VL tasks. CoOp \cite{zhou2022learning} pioneered prompt learning for CLIP by optimizing continuous prompt vectors in its language branch for few-shot image recognition. CoCoOp \cite{zhou2022conditional} extends CoOp by improving generalization through conditioning text prompts on visual features. MaPLe \cite{khattak2023maple} further advances this by introducing a multi-modal prompt tuning framework that enhances transferability through the joint learning of hierarchical prompts across both the vision and language branches. Despite recent advances, prompt learning approaches still struggle with overfitting. To address this, we fine-tune learnable prompts using a novel consistency-based alignment that incorporates an innovative style shift learning approach, enhancing generalizability without relying on external data.

\subsection{Consistency Regularization}
Regularization techniques, including weight decay \cite{loshchilov2017decoupled}, dropout \cite{srivastava2014dropout}, and data augmentation\cite{choi2022tokenmixup} are widely employed to prevent overfitting and enhance generalization to new data \cite{lee2022cross}. A recent study, CoPrompt\cite{roy2023consistency}, proposes a consistency regularization technique to improve the generalizability of the model by aligning the features between the prompted and frozen CLIP. However, this approach fails to account for the essential synergy between the visual and textual domains. Our approach addresses this by not only aligning feature-based consistency but also the cross-modal outputs of the prompted and frozen CLIP, strengthening the interaction between text and vision encoders. Additionally, we introduce a style shift learning technique instead of image-level perturbations \cite{roy2023consistency}, enhancing generalization capability across diverse styles without incurring computational overhead.


\section{Method}
\label{sec:formatting}
\subsection{Preliminaries}\label{sec:CLIP_}
\noindent\textbf{{Contrastive Language-Image Pre-training (CLIP).}} CLIP \cite{radford2021learning} has demonstrated a robust capability to learn open-set visual concepts. This model utilizes visual and textual encoders to produce corresponding embeddings from a given image and its associated textual description. Let $f(.)$ and $g(.)$ represent the image encoder and text encoder of CLIP, respectively. Their pre-trained parameters are denoted as $\theta_{CLIP} = {\theta_{f},\theta_{g}}$, where $\theta_{f}$ refers to the image encoder parameters and $\theta_{g}$ refers to the text encoder parameters. For a given image $X$, the extraction of image features begins by dividing the image into ${\mathbf{P}}^{2}$ patches. These patches are then processed through a projection operation known as $\operatorname{Patch Embed}$, which splits the input image into fixed-size patches and projects them into the feature space. Subsequently, a learnable class token $CLS$ is concatenated with these features, forming the sequence $\tilde{X} = \{CLS, e_{1}, e_{2}, \dots, e_{{\mathbf{P}}^{2}}\}$. This sequence is then passed through $L$ transformer layers to generate the visual feature representation, $\tilde{\mathbf{f}} \in \mathbb{R}^d$. Similarly, given the class name of the $i^{th}$ class, denoted as $y$, the $\operatorname{Word Embed}$ component first transforms the hand-crafted description, such as "a photo of a {class name}," into a sequence of vectorized textual tokens. This sequence can be represented as $\tilde {Y} = \{t_{SOS}, t_{1}, t_{2},..., t_{T},c_{k},t_{EOS}\}$, where $c_{k}$ is the word embedding corresponding to the class label, and $t_{SOS}$ and $t_{EOS}$ are learnable start and end tokens, respectively.  The text encoder $g$ encodes $\tilde {Y}$ via multiple transformer layers to produce the latent text feature, $\tilde{\mathbf{g}} \in \mathbb{R}^d$. For zero-shot inference, text features of the text template with class labels $\{1, 2,..., C\}$ are matched with the image feature as 
$\frac{\exp(\operatorname{sim}(\tilde{\mathbf{g}}_i, \tilde{\mathbf{f}}) / \tau)}{\sum_{i=1}^C \exp(\operatorname{sim}(\tilde{\mathbf{g}}_i, \tilde{\mathbf{f}}) / \tau)}$, where $\operatorname{sim}()$ denotes the cosine similarity and $\tau$ is the temperature.

\begin{figure*}[t]

    \centering 
    \includegraphics[scale=0.587]{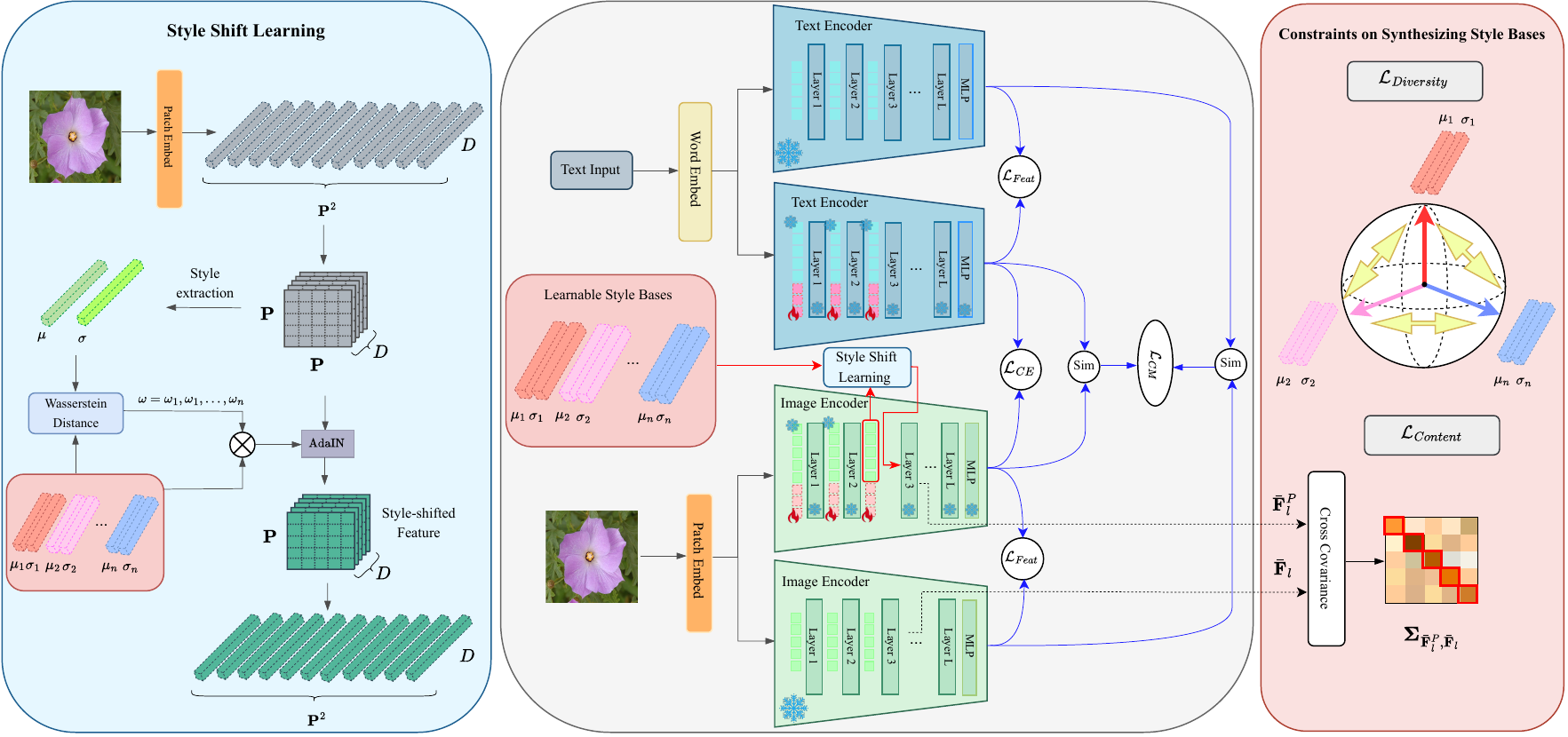}
    \caption{Overview of the proposed Style-Pro framework. Style-Pro introduces a style-guided prompt learning framework, incorporating a novel style shift learning approach in the feature space through a learnable set of style bases. Unseen styles are mapped into the style representation space as a weighted combination of style bases, reducing style discrepancies and improving performance on OOD data. Furthermore, Style-Pro ensures consistency between the embeddings of the prompted and frozen models during adaptation, which facilitates fine-tuning of CLIP while preserving its generalization capabilities.}
    \label{fig:Proposed Network}%
\end{figure*}

\vspace{2mm}
\noindent\textbf{{Prompt Learning.}} Inspired by prompt learning in NLP, numerous studies have explored the VL models by incorporating learnable prompt tokens during end-to-end training. In this study, we utilize hierarchical learnable prompt tokens independently for the text and image encoders, following the simple baseline method called Independent Vision-Language Prompting (IVLP) \cite{rasheed2023fine}. 
We concatenate learnable language prompts, denoted as $P_t = \left\{ p_t^1, p_t^2, \cdots, p_t^T \right\}$, and visual prompts, denoted as $P_v = \left\{ p_v^1, p_v^2, \cdots, p_v^V \right\}$, into the respective sets of textual and visual input tokens. Thus, the image encoder generates the prompted visual feature $\tilde{\mathbf{f}}_p = f(\tilde{X}_p, \theta_f)$ from input tokens $\tilde{X}p = \{P_v, e_{CLS}, e_1, e_2, \cdots, e_{\mathbf{P}^{2}}\}$, and the textual feature $\tilde{\mathbf{g}}_p = g(\tilde{Y}p, \theta_g)$ is obtained from $\tilde{Y}p = \{t_{SOS}, P_t, t_1, t_2, \cdots, t_T, c_k, t_{EOS}\}$. It should be noted that in this work, we employ deep prompting, which involves learning distinct sets of prompts for each transformer layer.
For image classification on a downstream dataset $D_s$, prompts interact with frozen $\theta_f$ and $\theta_g$ and are optimized with the cross-entropy loss as:
\begin{equation}
\mathcal{L}_{CE} = \frac{1}{N} \sum_{(X,y) \in \mathcal{D}_s} \frac{\exp \left( \operatorname{sim}(\tilde{\mathbf{f}}_p, \tilde{\mathbf{g}}_y) / \tau \right)}{\sum_{i=1}^{C} \exp \left( \operatorname{sim}(\tilde{\mathbf{f}}_p, \tilde{\mathbf{g}}_i) / \tau \right)}. 
\end{equation}
\subsection{Style-Guided Prompt Learning}
Fig. \ref{fig:Proposed Network} presents an overview of the Style-Pro framework. This framework integrates two complementary strategies: style shift learning and consistency constraints. Style-Pro incorporates an innovative style shift learning strategy to synthesize diverse style shifts and enhance robustness against OOD data. In parallel, Style-Pro introduces consistency constraints to preserve the alignment of the embeddings between the style-shifted prompted model and the pre-trained frozen CLIP, minimizing deviation during adaptation to downstream tasks.
\vspace{3mm}\\
\noindent \textbf{Style Shift Learning.} Adapting CLIP to downstream tasks often compromises OOD performance, as prompt learning is restricted to the training data distribution. To tackle the challenge of handling arbitrary unseen domains during testing, it is crucial to reduce the style discrepancies between training and test distributions. Our framework introduces a novel style shift learning approach using a learnable set of style bases, which addresses this issue more effectively than traditional image-based augmentation methods \cite{cubuk2020randaugment, hong2021stylemix}. This approach is computationally efficient and covers a broader range of styles beyond the source distribution without additional images.

Several studies \cite{zhou2021domain, huang2017arbitrary} have demonstrated that the shallow feature distribution of networks can reflect the style information of the input image. For a feature map $\mathbf{F} \in \mathbb{R}^{h \times w \times c}$, AdaIN \cite{huang2017arbitrary} shows that the channel-wise mean $\mu \in \mathbb{R}^c$ and standard deviation $\sigma \in \mathbb{R}^c$ convey information about the style of the input image. Therefore, stylizing a given feature $\mathbf{F}$ with an arbitrary target style ($\mu_t, \sigma_t$) is performed as follows:
\begin{equation}\label{adain_equ}
\text{AdaIN}(\mathbf{F}, \mu_t, \sigma_t) = \sigma_t \left( \frac{\mathbf{F}- \mu(\mathbf{F})}{\sigma(\mathbf{F})} \right) + \mu_t,
\end{equation}
where $\mu(\cdot)$ and $\sigma(\cdot)$ denote the mean and standard deviation of the input feature, respectively.

Given an input image, as discussed in Section \ref{sec:CLIP_}, the \( l \)-th layer of ViT generates ${\mathbf{P}}^{2}+1$ token representations. Among these, ${\mathbf{P}}^{2}$ representations are derived from the image patches, and one of them corresponds to the learnable classification token $(\mathbf{F}_l \in \mathbb{R}^{({{\mathbf{P}}^{2}}+1) \times D})$, where $D$ represents the dimension of each token representation. To extract the style information from the \( l \)-th layer of ViT, following the methodology established by \cite{fu2023styleadv}, we employ AdaIN across the dimension of the token representations. As illustrated in Fig. \ref{fig:Proposed Network}, this process involves reshaping the patch representations into $\mathbf{F}'_l \in \mathbb{R}^{\mathbf{P} \times \mathbf{P} \times D}$. Subsequently, we can calculate the style information for the token representations as follows;
\begin{equation}
\mu(\mathbf{F}'_{l})= \frac{1}{{\mathbf{P}}^{2}} \sum_{k=1}^{\mathbf{P}} \sum_{k=1}^{\mathbf{P}} \mathbf{F}'_{l},
\end{equation}
\begin{equation}
\sigma(\mathbf{F}'_{l})= \sqrt{\frac{1}{{\mathbf{P}}^{2}} \sum_{k=1}^{\mathbf{P}} \sum_{k=1}^{\mathbf{P}} \left( \mathbf{F}'_{l} - \mu(\mathbf{F}'_l) \right)^2 + \epsilon},
\end{equation}
\noindent where $\mu(\mathbf{F}'_{l}), \sigma(\mathbf{F}'_l) \in \mathbb{R}^D$. Considering a collection of style bases $B_{sty} = {(\mu_b^n, \sigma_b^n)}_{n=1}^{N}$, we enhance the styles of training features to improve the generalization ability to new domains. Initially, we compute the Wasserstein distance \cite{vallender1974calculation} to determine the discrepancy in style distribution between the current image ($\mu_{cur}, \sigma_{cur}$) and the \( n \)-th style basis $(\mu_b^{n}, \sigma_b^{n})$ as follows:
\begin{equation}
d_{cur} = ||\mu_{cur} - \mu_b^n||_2^2 + (\sigma_{cur}^2 + {\sigma_b^{n}}^2 - 2\sigma_{cur}\sigma_b^n).
\end{equation}

Next, the reciprocal of $d_{cur}$ is used to measure the similarity between the current image and the \( n \)-th style basis:
\begin{equation}
\omega_n = \frac{\exp(1/(1 + d_{cur}))}{\sum_{n=1}^{N} \exp(1/(1 + d_{cur}))},
\end{equation}
where the softmax function ensures that the sum of $\omega = {\omega_n | n = 1, 2, \ldots, N }$ equals 1. Based on the estimated similarity $\omega$, the mapped style $(\mu_{map}, \sigma_{map})$ is obtained by the weighted sum of the style bases:
\begin{equation}
\mu_{map} = \sum_{n=1}^{N} \omega_n \mu_b^n, \quad \sigma_{map} = \sum_{n=1}^{N} \omega_n \sigma_b^n.
\end{equation}
Therefore, the style bases most similar to the unseen style play a more dominant role in mapping the unseen style. Finally, using Eq. \ref{adain_equ}, the mapped style is applied to the normalized feature ($\bar{\mathbf{F}'_l}$) to obtain the style-shifted feature (${\mathbf{F}}_{l}''$) as follows:
\begin{equation}
{\mathbf{F}_{l}''} = \sigma_{map} \bar{\mathbf{F}'_l} + \mu_{map}.
\end{equation}

Following this operation, the style-shifted token representations are fed into the subsequent layer of the prompted encoder.

\vspace{2mm}
\noindent \textbf{Constraints on Synthesizing Styles.} To address the various domain shifts, our proposed method adheres to two constraints to produce optimized and efficient style bases. The first constraint ensures that the synthesized styles comprehensively cover the feature space, thereby enhancing the ability of the model to generalize across different domains. To achieve this, we employ a novel loss function that compels the model to generate a diverse set of styles, each representing different points on the hyper-sphere. To maximize the diversity of $N$ style bases in a hyper-spherical feature space, each style basis must be orthogonal to all other style bases. Consequently, we minimize the absolute value of the cosine similarity between the \( n \)-th style basis and every other existing style basis, as expressed by the following equation:
\begin{equation}
\mathcal{L}_{Diversity} = \sum_{\substack{n=1}}^{N}(\sum_{\substack{k=1 \\ k \neq n}}^{N} \left| \frac{\mu^{n}_{b}}{\| \mu^{n}_{b} \|} \cdot \frac{\mu^{k}_{b}}{\| \mu^{k}_{b} \|} \right| + \sum_{\substack{k=1 \\ k \neq n}}^{N} \left| \frac{\sigma^{n}_{b}}{\| \sigma^{n}_{b} \|} \cdot \frac{\sigma^{k}_{b}}{\| \sigma^{k}_{b} \|} \right|).
\end{equation}

Focusing only on the diversity constraint is insufficient for synthesizing appropriate style bases. It is important to have diverse styles that do not damage the content information of the features. This requires that in each layer, features from both the prompted and frozen vision encoders convey identical content. Imposing this constraint ensures that the model preserves content consistency when shifting styles in the prompted vision encoder. Given the normalized features, $\bar{\mathbf{F}}_{l}^P$ and $\bar{\mathbf{F}}_{l}$, from the \( l \)-th layer of the prompted and frozen vision encoders, respectively, the cross-covariance matrix is computed as follows:
\begin{equation}
\mathbf\Sigma_{\bar{\mathbf{F}}_{l}^P, \bar{\mathbf{F}}_{l}} = \mathbf{E}[\bar{\mathbf{F}}_{l}^P \cdot (\bar{\mathbf{F}}_{l})^T].
\end{equation}
Since $\bar{\mathbf{F}}_{l}^P$ and $ \bar{\mathbf{F}}_{l}$ should contain identical content, the diagonal elements of their cross-covariance matrix are expected to be equal to 1 \cite{zbontar2021barlow}. Consequently, we define the content consistency loss as follows:
\begin{equation}
\mathcal{L}_{Content} = \| \operatorname{diag}(\mathbf\Sigma_{\bar{\mathbf{F}}_{l}^P, \bar{\mathbf{F}}_{l}}) - \mathbf{1} \|_2, 
\end{equation}
where \(\operatorname{diag}() \in \mathbb{R}^D\) denotes the column vector comprising diagonal elements of \(\mathbf\Sigma_{{\bar{\mathbf{F}}_{l}^P, \bar{\mathbf{F}}_{l}}}\) and \(\mathbf{1} \in \mathbb{R}^D\) denotes the one vector.
\vspace{2mm}\\
\noindent\textbf{Self-Consistency Regularization.}
As outlined in the previous subsection, merely training prompts with a supervised task-specific loss does not adequately preserve the general attributes inherent in the frozen CLIP. Consequently, despite keeping the weights of the CLIP image and text encoders fixed, their performance on new tasks deteriorates. To overcome this limitation, we propose a novel regularization method that ensures the prompted model maintains its generalization capability across new classes and diverse domains. This is accomplished by applying complementary constraints, which allow CLIP to retain its general knowledge while learning prompts for downstream tasks.

To effectively leverage the guidance of the frozen model, it is crucial to minimize the variations introduced by the prompts across different modality branches. To this end, we impose two distinct constraints on the prompted visual and textual features to ensure their alignment with the frozen CLIP features. For a given input sample and its corresponding textual label, we extract textual and visual features using both the prompted and frozen latent spaces. Subsequently, we apply the following two complementary loss functions:

\begingroup
    \setlength{\leftmargini}{1.8pt}
    \setlength{\labelsep}{1.9 pt}
 \begin{itemize}[align=left,leftmargin=*]
\item \textbf{Feature-level Alignment.} To evaluate the similarity between feature representations, we utilize the Mean Square Error (MSE) as a metric to quantify the average squared differences. This approach accurately assesses how closely the predicted embeddings align with the frozen model embeddings. The loss function is defined as:
\begin{equation}
{\mathcal{L}_{Feat}} = \frac{1}{d} \left(  \lambda_f \sum_{i=1}^{d} (\tilde{\mathbf{f}}_i - \tilde{\mathbf{f}}_{p_{i}})^2 +\lambda_g \sum_{i=1}^{d} \left(\tilde{\mathbf{g}}_i - \tilde{\mathbf{g}}_{p_{i}} \right)^2\right),
\end{equation}
where $\tilde{\mathbf{f}}_{p_{i}}$ and $\tilde{\mathbf{g}}_{p_{i}} \in \mathbb{R}^d $ represent the vision and text embeddings derived from the prompted model, respectively. The parameters $\lambda_f$ and $\lambda_g$ are weighting factors that control the contribution of the vision and text modalities to the overall loss. 
\item \textbf{Cross-modality Alignment.} The performance of VL models can degrade substantially when there is a misalignment between the visual and textual domains. These discrepancies often arise due to inherent differences in how visual and textual data are represented and processed within these models. To tackle this challenge and enhance the integration between the text and vision-prompted encoders, we further align the output predictions of the prompted model ($Pre_p$) with those of the pre-trained frozen CLIP ($Pre$) by employing Kullback-Leibler (KL) divergence. Specifically, we define the cross-modality alignment loss as:
\begin{equation}
\mathcal{L}_{CM} = \mathcal{D}_{\text{KL}}(Pre,Pre_p ),
\end{equation}
\begin{equation}
Pre =  \operatorname{sim}(\tilde{f}, \tilde{g}), \: Pre_p =  \operatorname{sim}(\tilde{f}_p, \tilde{g}_p),
\end{equation}
where the function sim() represents cosine similarity. 
This alignment enables the model to achieve a comprehensive understanding of both visual and textual information. As a result, our method enhances the generalization performance across diverse domains, mitigating the risk of performance degradation due to discrepancies between the text and vision modalities.
\end{itemize}
\endgroup

\vspace{-2mm}
We utilize the weighted combination of all the mentioned loss terms for the end-to-end training. Therefore, our final loss function is:
\begin{multline}
\mathcal{L}_{totall} = \mathcal{L}_{CE} +
{\mathcal{L}_{CM}}+ {\mathcal{L}_{Feat}} + \\
\lambda_1 {\mathcal{L}_{Diversity}} + \lambda_2 {\mathcal{L}_{Content}},
\end{multline}
\noindent where $\lambda_1$ and $\lambda_2$ are hyperparameters that control the relative importance of the diversity and content losses, respectively.

\section{Experiments}
\subsection{Datasets and Implementation Details }
Following previous prompt tuning studies \cite{rasheed2023fine, khattak2023maple,zhou2022conditional}, we validate our method across three different settings: generalization from base-to-novel classes, cross-dataset evaluation, and domain generalization. Our experiments incorporate a variety of datasets, including two generic-object datasets (ImageNet \cite{deng2009imagenet} and Caltech101 \cite{fei2004learning}), fine-grained datasets (OxfordPets \cite{parkhi2012cats}, StanfordCars \cite{krause20133d}, Flowers102 \cite{nilsback2008automated}, Food101 \cite{bossard2014food}, and FGVCAircraft\cite{maji2013fine}), a remote sensing classification dataset (EuroSAT\cite{helber2019eurosat}), a scene recognition dataset (SUN397 \cite{xiao2010sun}), an action recognition dataset (UCF101\cite{soomro2012ucf101}), and a texture dataset (DTD\cite{cimpoi2014describing}). For domain generalization, we employ ImageNet as the source dataset, and its four variants as target datasets, including ImageNetV2 \cite{recht2019imagenet}, ImageNetSketch \cite{wang2019learning}, ImageNet-A\cite{hendrycks2021natural}, and ImageNet-R \cite{hendrycks2021many}. All these experiments are conducted using a 16-shot setting, meaning that only 16 training examples are provided for each category.

In our experiments, we utilize a ViT-B/16-based CLIP model, employing deep prompting with \(V = T = 4\) VL prompts. It should be noted that the prompts are randomly initialized using a normal distribution, except for the text prompts in the first layer, which are initialized with the word embeddings of ``a photo of a". For domain generalization and cross-dataset evaluation, learnable prompts are injected into the first three transformer layers. In the base-to-novel setting, prompts are injected into the first nine transformer layers. Each model is trained for 25 epochs with a batch size of 4 and a learning rate of 0.0025 using the SGD optimizer on a single NVIDIA RTX 4090 GPU. The results are averaged over three runs. We set the parameters $\lambda_f$ and $\lambda_g$ in $\mathcal{L}_{MSE}$ to 15 and 25, respectively. The diversity loss $\mathcal{L}_{diversity}$ and the content loss $\mathcal{L}_{content}$ are weighted by factors of 0.005 and 0.2, respectively. The number of style bases, $N$, is fixed at 12 for all experiments.

\subsection{Base-to-Novel Generalization}

To evaluate the generalization capability of our method within a dataset, we conduct evaluations on 11 different image classification datasets. Adhering to the experimental setup established by previous studies \cite{zhou2022conditional, khattak2023maple, lu2022prompt}, each dataset is divided into base and novel classes. The model is trained on the base classes utilizing a few-shot learning approach with 16 shots and is subsequently evaluated on both the base and novel classes. 
The results, presented in Table \ref{tab:base-to-novel}, compare our proposed method with state-of-the-art techniques, including Zero-shot CLIP \cite{radford2021learning}, CoOp \cite{zhou2022learning}, CoCoOp \cite{zhou2022conditional}, MaPLe \cite{khattak2023maple}, CoPrompt \cite{roy2023consistency}, PromptSRC \cite{khattak2023self}, and MMA \cite{yang2024mma}, across the 11 datasets.
Notably, Style-Pro demonstrates exceptional performance in addressing one of the most challenging categories, unseen classes, underscoring its superior zero-shot generalizability to novel scenarios. Specifically, our Style-Pro achieves a 0.83\% average gain over CoPrompt \cite{roy2023consistency} on novel class generalization. It is also important to highlight that while Style-Pro surpasses its competitors in handling new classes, it maintains effective adaptation to base classes with an average 0.22\% improvement over the second-best result. Although a few methods marginally outperform Style-Pro in specific base classes, the difference is negligible. These results highlight the robustness of Style-Pro in adapting to downstream tasks while still maintaining strong generalization ability.

\subsection{Cross-Dataset Evaluation}

In cross-dataset evaluation, similar to the state-of-the-art methods \cite{zhou2022conditional, lu2022prompt, roy2023consistency}, we train the model on ImageNet \cite{deng2009imagenet} and directly evaluate it on other datasets without any data-specific fine-tuning. Table \ref{tab:cross_comparison} summarizes the results of our cross-dataset evaluation. Style-Pro demonstrates an average improvement of 0.24\% over existing methods, consistently outperforming them across various datasets. This strong performance highlights the good zero-shot transferability of the proposed Style-Pro, making it highly suitable for applications requiring adaptability across diverse datasets.

\subsection{Domain Generalization}

Table \ref{tab:domain} presents a summary of the results for Style-Pro in comparison with previous methods on OOD datasets. In line with prior works \cite{zhou2022conditional, lu2022prompt, roy2023consistency, yang2024mma, roy2023consistency}, we fine-tune our model on ImageNet dataset \cite{deng2009imagenet} and evaluate it on several ImageNet variants. As shown in Table \ref{tab:domain} , our approach consistently outperforms existing methods in terms of robustness to domain shifts within the domain generalization setting. This demonstrates that style shift learning enables the model to map the styles of unknown domains into a dedicated style representation space, thus minimizing discrepancies in style and improving performance on OOD datasets.

\begin{table}[t]
\centering
  \setlength\tabcolsep{2pt}
\resizebox{0.485\textwidth}{!}{\begin{tabular}{lcccccccccc}

 \hline \hline
\addlinespace[1mm]
\multirow{2}{*}{\textbf{Dataset}} &  & \textbf{CLIP} & \textbf{CoOp} & \textbf{CoCoOp} & \textbf{MaPLe} & \textbf{PromptSRC} &\textbf{CoPrompt}&\textbf{MMA}&\textbf{Style-Pro} \\
{} &  & \cite{radford2021learning} & \cite{zhou2022learning} & \cite{zhou2022conditional} & \cite{khattak2023maple}   & \cite{khattak2023self} &\cite{roy2023consistency}& \cite{yang2024mma}&(Proposed)  \\
\midrule

\multirow{2}{*}{\textbf{Average on}} &{B}& {69.34} & {82.69} & {80.47}& {82.28} & {84.26} & {84.00}& 83.20& \textbf{84.48} \\
&{N} & {74.22} & {63.22} & {71.69}& {75.14} &{76.10} & {77.23}&76.80& \textbf{78.06} \\
\textbf{11 datasets} &{H} & {71.70} & {71.66} & {75.83} & {78.55} &{79.97}& {80.48} &79.87&\textbf{80.98}\\

\midrule
\multirow{3}{*}{\textbf{ImageNet}} & {B}& 72.43 & 76.47 & 75.98 & 76.66 & 77.60 & \textbf{77.67}&77.31& {77.58} \\
 & {N}& 68.14 & 67.88 & 70.43 & 70.54 &70.73 & {71.27} &71.00& \textbf{71.68}\\

 & H& 70.22 & 71.92 & 73.10 & 73.47 &  74.01 & {74.33}&74.02& \textbf{74.51}\\
\midrule
\multirow{3}{*}{\textbf{Caltech101}} & B& 96.84 & 98.00 & 97.96 & 97.74 & 98.10 & {98.27}& \textbf{98.40}& 98.38\\
& N& 94.00 & 89.81 & 93.81 & 94.36 & 94.03 & {94.90}&  94.00& \textbf{95.44}\\

& H& 95.40 & 93.73 & 95.84 & 96.02 &  96.02& {96.55} & 96.15&\textbf {96.89} \\
\midrule
\multirow{3}{*}{\textbf{OxfordPets}} & B&91.17 & 93.67 & 95.20 & 95.43&95.33& \textbf{95.67} & 95.40& 95.64\\
& N & 97.26 & 95.29 & 97.69 & 97.76 &  97.30 & {98.10}&  98.07& \textbf{98.63}\\

 &H& 94.12 & 94.47 & 96.43 & 96.58 & 96.30 & {96.87}& 96.72& \textbf{97.11}\\
\midrule
\multirow{2}{*}{\textbf{Stanford}} & B& 63.37 & 78.12 & 70.49 & 72.94 & 78.27 & {76.97}& 78.50& \textbf{78.53}\\
 &N&  74.89 & 60.40 & 73.59 & 74.00 &  74.97 & {74.40}& 73.10& \textbf{75.12}\\

\textbf{Cars} & H& 68.65 & 68.13 & 72.01 & 73.47 & 76.58 & {75.66}&75.70& \textbf{76.79} \\
\midrule
\multirow{2}{*}{\textbf{Flowers}} & B& 72.08 & 97.60 & 94.87 & 95.92 & \textbf{98.07} & {97.27}& 97.77& 98.04 \\
 & N& 77.80 & 59.67 & 71.75 & 72.46 &  {76.50} &{76.60}&75.93& \textbf{76.86}\\
{\textbf{102}}&H& 74.83 & 74.06 & 81.71  & 82.56 &  85.95 & {85.71}& 85.48& \textbf{86.17} \\
\midrule
\multirow{3}{*}{\textbf{Food101}} &B& 92.43 & 88.33 & 90.70  & 90.71 & 90.67& {90.73} &90.13& \textbf{90.93}\\
 &N&91.22 & 82.26& 91.29 & 92.05 &  91.53 & {92.07}&91.30& \textbf{92.29}\\

 & H& 90.66 & 85.19 & 90.99 & 91.38 &  91.10 & {91.40}& 90.71& \textbf{91.60} \\
\midrule
\multirow{2}{*}{\textbf{FGVC}}  & B& 27.19 & 40.44 & 33.41  & 37.44 &  42.73 & {40.20}& 40.57& \textbf{42.79} \\
& N& 36.29 & 22.30 & 23.71  & 35.61 &  37.87 & \textbf{39.33}&36.33 & 39.28\\

 {\textbf{Aircraft}}& H& 31.09 & 28.75 & 27.74 &  36.50 & 40.15 & {39.76}& 38.33& \textbf{40.96}\\
\midrule
\multirow{3}{*}{\textbf{SUN397}}&B & 69.36 & 80.60 & 79.74& 80.82 &  \textbf{82.67}& {82.63} &82.27& 82.66\\
 & N&75.35 & 65.89 & 76.86 & 78.70 &  78.47 & {80.03}& 78.57& \textbf{80.61}\\

 & H&72.23 & 72.51 & 78.27 & 79.75 &  80.52 & {81.31}& 80.38& \textbf{81.62}\\
\midrule
\multirow{3}{*}{\textbf{DTD}} & B&53.24 & 79.44 & 77.01 & 80.36 & 83.37 & {83.13}&83.20& \textbf{83.41} \\
 & N& 59.90 & 41.18 & 56.00 & 59.18 &  62.97 & {64.73}& \textbf{65.63}& 65.58\\

 & H& 56.37 & 54.24 & 64.85 & 68.16 &  71.75 & {72.79}& 73.38& \textbf{73.43}\\
\midrule
\multirow{3}{*}{\textbf{EuroSAT}} & B& 56.48 & 92.19 & 87.49 & 94.07 & 92.90 & \textbf{94.60} &85.46& 94.52\\
 & N& 64.05 & 54.74 & 60.04 & 73.23 & 73.90 & 78.57 & 82.34&\textbf{82.74} \\

 & H& 60.03 & 68.69 & 71.21 & 82.35 &  82.32 & {85.84}& 83.87& \textbf{88.24}\\
\midrule
\multirow{3}{*}{\textbf{UCF101}} &B& 70.53 & 84.69 & 82.33 & 83.00 & 87.10 & \textbf{86.90}&86.23& 86.83 \\
 & N& 77.50 & 56.05 & 73.45  & 78.66 &  78.80 &{79.57} &80.03& \textbf{80.40}\\

 & H& 73.85 & 67.46 & 77.64 & 80.77 &  82.74 & {83.07} &82.20& \textbf{83.49}\\
\hline \hline
\end{tabular}}
\caption{Comparison of our proposed method with state-of-the-art methods on different datasets in the base-to-novel generalization setting. `B' and `N' denote accuracy on base and novel classes, respectively. `H' represents the harmonic mean of base and novel accuracy, reflecting the balance between adaptation and generalization.}
\label{tab:base-to-novel}
\end{table}

\begin{table}[!]
    \centering
    \footnotesize
       \setlength\tabcolsep{3.5pt}

    \begin{tabular}{lccccccc}
          \hline \hline
    \addlinespace[0.5mm]
        & \multicolumn{1}{c}{\textbf{Source}} & \multicolumn{5}{c}{\textbf{Target}} & \\
        \cmidrule(lr){2-2} \cmidrule(lr){3-7}
        & \textbf{ImageNet} & \textbf{-V2} & \textbf{-S} & \textbf{-A} & \textbf{-R} & \textbf{Avg.} \\
        \midrule
        \textbf{CLIP} \cite{radford2021learning}& 66.73 & 60.83 & 46.15 & 47.77 & 73.96 & 57.18 \\
        \textbf{CoOp} \cite{zhou2022learning}& 71.51 & 64.20 & 47.99 & 49.71 & 75.21 & 59.28 \\
        \textbf{CoCoOp} \cite{zhou2022conditional}& 71.02 & 64.07 & 48.75 & 50.63 & 76.18 & 59.91 \\
        \textbf{MaPLe}  \cite{khattak2023maple}& 70.72 & 64.07 & 49.15 & 50.90 & 76.98 & 60.27 \\
        \textbf{PromptSRC} \cite{khattak2023self}& \textbf{71.27} & 64.35 & 49.55 & 50.90 & 77.80 & 60.65 \\
        \textbf {CoPrompt} \cite{roy2023consistency}& 70.80 & 64.81 & 49.54 & 51.51 & 77.34 & 60.80 \\
          \textbf{MMA} \cite{yang2024mma}& 71.00 & 64.33 & 49.13 & 51.12 & 77.32 & 60.77 \\
        \hdashline [2pt/1.5pt]
         \addlinespace[0.5mm]
       
          \textbf{Style-Pro}& {71.23} & \textbf{65.66} & \textbf{50.38} & \textbf{51.93} & \textbf{77.98} & \textbf{61.49} \\
          \hline \hline
    \end{tabular}
    \caption{Comparison of our proposed method with state-of-the-art studies in
the domain generalization setting.}
    \label{tab:domain}
\end{table}

\begin{table*}[h!]
    \centering
    \footnotesize
   \setlength\tabcolsep{4.4pt}
    \begin{tabular}{lcccccccccccc}
    \hline \hline
    \addlinespace[0.5mm]
    & \multicolumn{1}{c}{\textbf{Source}} & \multicolumn{11}{c}{\textbf{Target}} \\
    \cmidrule(lr){2-2} \cmidrule(lr){3-13}
    & \rotatebox{45}{\textbf{ImageNet}} & \rotatebox{45}{\textbf{Caltech101}} & \rotatebox{45}{\textbf{OxfordPets}} & \rotatebox{45}{\textbf{StanfordCars}} & \rotatebox{45}{\textbf{Flowers102}} & \rotatebox{45}{\textbf{Food101}} & \rotatebox{45}{\textbf{Aircraft}} & \rotatebox{45}{\textbf{SUN397}} & \rotatebox{45}{\textbf{DTD}} & \rotatebox{45}{\textbf{EuroSAT}} & \rotatebox{45}{\textbf{UCF101}} & \rotatebox{45}{\textbf{Average}} \\
    \midrule
    \textbf{CoOp} \cite{zhou2022learning}& 71.51 & 93.70 & 89.14 & 64.51 & 68.71 & 85.30 & 18.47 & 64.15 & 41.92 & 46.39 & 66.55 & 63.88 \\
    \textbf{CoCoOp} \cite{zhou2022conditional}& 71.02 & 94.43 & 90.14 & 65.32 & 71.88 & 86.06 & 22.94 & 67.36 & 45.73 & 45.37 & 68.21 & 65.74 \\
    
    \textbf{MaPLe} \cite{khattak2023maple}& 70.72 & 93.53 & 90.49 & 65.57& 72.23& 86.20 & 24.74 & 67.01 & 46.49 & 48.06 & 68.69 & 66.30 \\

      \textbf{PromtSCR} \cite{khattak2023self}& \textbf{71.27} &93.60 &90.25& 65.70& 70.25 &86.15& 23.90 &67.10 &46.87& 45.50 &68.75 &65.81 \\

   \textbf{CoPrompt} \cite{roy2023consistency}& 70.80& 94.50 &90.73& 65.67 &72.30 &86.43& 24.00& 67.57&47.07& \textbf{51.90}& 69.73 &67.00\\
      \textbf{MMA} \cite{yang2024mma}&71.00 &93.80 &90.30& \textbf{66.13} &72.07 &86.12& \textbf{25.33} &68.17& 46.57& 49.24& 68.32& 66.61\\ 
 
      \hdashline [2pt/1.5pt]

      \addlinespace[0.5mm]
   \textbf{Style-Pro}& 71.23 & \textbf{94.66}& \textbf{90.91} & 66.03 & \textbf{72.54} & \textbf{86.61} & 25.14 & \textbf{68.38} & \textbf{47.29} & 50.85 & \textbf{69.96} & \textbf{67.24} \\
      \hline \hline
  
    \end{tabular}
    \caption{Comparison of our proposed method with state-of-the-art approaches in cross-dataset evaluation. Our method achieves superior average performance across 10 datasets, highlighting its strong zero-shot adaptability.}
\label{tab:cross_comparison}
\end{table*}

\begin{table}[h!]
    \centering
    \footnotesize
       \setlength\tabcolsep{5.9pt}

    \begin{tabular}{l c c c c| c c c}
          \hline \hline
        & \multicolumn{4}{c|}{\textbf{Approach}} & \multicolumn{3}{c}{\textbf{Accuracy}} \\
        \hline
        & \multicolumn{2}{c|}{\textbf{Consistency}}&\multicolumn{2}{c|}{\textbf{Style Shift}} & \multirow {2}{*} {\textbf{Base}} & \multirow {2}{*}{\textbf{Novel}} &  \multirow {2}{*}{\textbf{HM}} \\
   \cline {1-5 }  & Feat & \multicolumn{1}{c|}{CM} &Content &  Diversity&  &  & \\
        \hline
           & & & & &82.51 &73.36 &77.66 \\
         & $\checkmark$ &  &  & & 82.77 & 74.28 &78.30\\
         & $\checkmark$  & $\checkmark$  &   &  & 82.97 & 75.64 & 79.14\\
       & $\checkmark$  & $\checkmark$  &$\checkmark$ &   &  83.11& 76.09  &79.45\\
        & $\checkmark$  & $\checkmark$  &$\checkmark$ & $\checkmark$ & \textbf{84.48} & \textbf{78.06} & \textbf{80.98}\\
          \hline \hline
    \end{tabular}
    \caption{ Analysis of different constraints of Style-Pro framework.}
    \label{tab:ablation}
\end{table}

\subsection{Ablation and Analysis}
\begin{figure}[!]

    \centering 
    \includegraphics[scale=0.32]{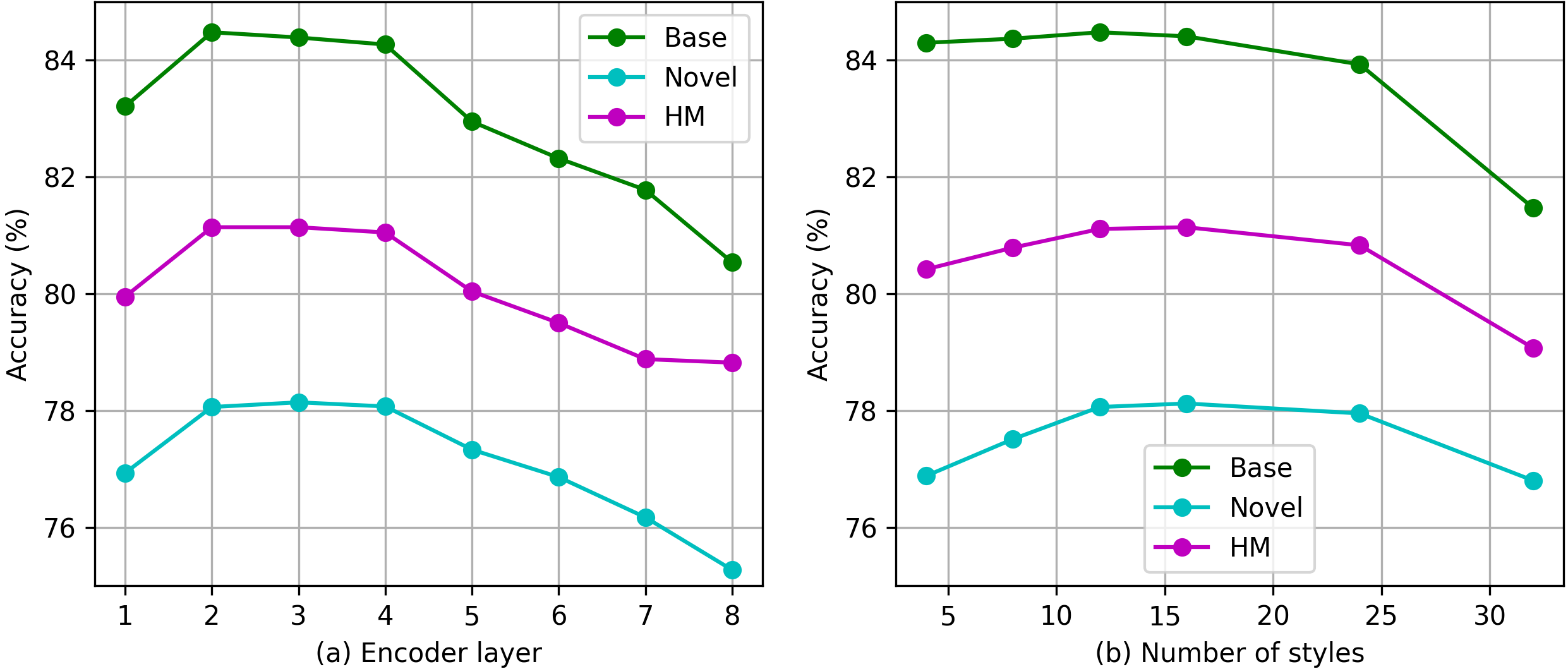}
    \caption{(a) Ablation study on style shift learning at different layers of the vision encoder. (b) Ablation study on the impact of the number of learnable style bases.}
    \label{fig:ablation}%
\end{figure}

In this subsection, we assess the individual contributions of each component of the proposed Style-Pro framework. We demonstrate how these components complement each other to mitigate overfitting in VL model adaptation, leading to improved generalization performance.\vspace{1mm}\\
\noindent\textbf{Effectiveness of each constraint.} Table \ref{tab:ablation} summarizes the results of the ablation study on the various constraints introduced in our Style-Pro framework, with the outcomes averaged over 11 datasets. To elucidate the impact of our alignment constraints, we conduct two distinct experiments. The first experiment considers only the feature-level constraint (second row), while the second incorporates both the feature-level and the cross-modality alignment (third row). We compare the performance of these approaches with the baseline approach, which utilizes only IVLP (first row). The results demonstrate that applying both constraints together enhances accuracy for both base and new classes, thereby confirming its effectiveness in improving generalizability. Additionally, we implement two further experiments to assess the impact of our style shift learning approach on the accuracy improvement of the VL model. The results (see Table \ref{tab:ablation}) indicate that incorporating content loss, which preserves the content information of style-shifted features, enhances the model's ability to discriminate effectively (as shown in the fourth row). Moreover, enforcing a broader range of synthetic style bases significantly boosts accuracy for both base and new classes (the fifth row), with a particularly notable improvement observed in the new classes, aligning with our primary objective.\vspace{1mm}\\
\noindent\textbf{Analysis of different augmentation approaches for the prompted vision encoder.} To further illustrate the impact of our proposed style shift learning approach, we compare its performance with alternative augmentation methods. As shown in Table \ref{tab:ablation22}, replacing our method with standard image-based augmentations, such as random resized cropping (second row), or a more advanced technique like RandAugment \cite{cubuk2020randaugment} (third row), does not yield significant improvements over the baseline model without any augmentation (first row). Furthermore, we explore the application of MixStyle \cite{zhou2021domain} as a substitute for our approach. Although MixStyle as a feature-level augmentation method demonstrates some performance gains, the results clearly indicate that our proposed approach substantially outperforms all other strategies. This highlights the superiority of our approach in boosting the generalization capacity of the model.\vspace{1mm}\\
\noindent\textbf{Analysis of the number of style bases and the layer selection for style shift learning.}
Fig. \ref{fig:ablation} (a) explores the effects of applying style shift learning at different layers of the vision encoder. The results indicate that applying style shift learning at the second layer achieves the optimal balance in accuracy across the base, novel, and HM categories, highlighting this layer as the most effective for integrating style variations.
Subsequently, we examine the influence of the number of learnable style bases, as shown in Fig. \ref{fig:ablation} (b). The curves reveal that a limited number of style bases fails to address the broad spectrum of domain shifts, resulting in poor generalization, while too many bases add unnecessary complexity, reducing performance. Thus, we set the number of style bases to 12 in all experiments.


 
       

\begin{table}[h]
    \centering
    \footnotesize

    \begin{tabular}{l| c}
          \hline \hline
         {\textbf{Approach}} & {\textbf{Accuracy}} (HM)\\

        \hline
      Baseline  &79.14\\
       Simple Augmentation & 79.81 \\
      RandAugment \cite{cubuk2020randaugment}& 79.75 \\
      MixStyle \cite{zhou2021domain}  & 79.93 \\
     Style-Pro   &\textbf{80.98}\\

          \hline \hline
    \end{tabular}
    \caption{Analysis of different augmentation approaches for the vision encoder.}
    \label{tab:ablation22}
\end{table}

\section{Conclusion}

In this work, we present a style-guided prompt learning framework for fine-tuning VL models in the context of few-shot image recognition. Our framework leverages a novel style shift learning approach that enhances the model's robustness to OOD data by utilizing learnable style bases. Further, we maintain alignment between the style-shifted prompted model and the frozen pre-trained CLIP, ensuring the frozen model's generalization capacity is preserved while fine-tuning for downstream tasks. Extensive experiments on 11 image classification benchmarks demonstrate that Style-Pro consistently outperforms existing methods, with more considerable gains in generalization to new classes and domains.

\section*{Acknowledgment}
This material is based upon work supported by the National Science Foundation under Grant Numbers CNS-2232048, and CNS-2204445.


{\small
\bibliographystyle{ieee_fullname}
\bibliography{egbib_edited}
}

\end{document}


\title{Style-Pro: Style-Guided Prompt Learning for Generalizable Vision-Language Models}

\author{First Author\\
Institution1\\
Institution1 address\\
{\tt\small firstauthor@i1.org}
\and
Second Author\\
Institution2\\
First line of institution2 address\\
{\tt\small secondauthor@i2.org}
}
\maketitle

\begin{abstract}
   The ABSTRACT is to be in fully justified italicized text, at the top of the left-hand column, below the author and affiliation information.
   Use the word ``Abstract'' as the title, in 12-point Times, boldface type, centered relative to the column, initially capitalized.
   The abstract is to be in 10-point, single-spaced type.
   Leave two blank lines after the Abstract, then begin the main text.
   Look at previous WACV abstracts to get a feel for style and length.
\end{abstract}

\section{Introduction}
\label{sec:intro}

Please follow the steps outlined below when submitting your manuscript to the IEEE Computer Society Press.
This style guide now has several important modifications (for example, you are no longer warned against the use of sticky tape to attach your artwork to the paper), so all authors should read this new version.

\subsection{Language}

All manuscripts must be in English.

\subsection{Dual submission}

Please refer to the author guidelines on the \confName\ \confYear\ web page for a
discussion of the policy on dual submissions.

\subsection{Paper length}
Papers, excluding the references section, must be no longer than eight pages in length.
The references section will not be included in the page count, and there is no limit on the length of the references section.
For example, a paper of eight pages with two pages of references would have a total length of 10 pages.
{\bf There will be no extra page charges for \confName\ \confYear.}

Overlength papers will simply not be reviewed.
This includes papers where the margins and formatting are deemed to have been significantly altered from those laid down by this style guide.
Note that this \LaTeX\ guide already sets figure captions and references in a smaller font.
The reason such papers will not be reviewed is that there is no provision for supervised revisions of manuscripts.
The reviewing process cannot determine the suitability of the paper for presentation in eight pages if it is reviewed in eleven.

\subsection{The ruler}
The \LaTeX\ style defines a printed ruler which should be present in the version submitted for review.
The ruler is provided in order that reviewers may comment on particular lines in the paper without circumlocution.
If you are preparing a document using a non-\LaTeX\ document preparation system, please arrange for an equivalent ruler to appear on the final output pages.
The presence or absence of the ruler should not change the appearance of any other content on the page.
The camera-ready copy should not contain a ruler.
(\LaTeX\ users may use options of wacv.sty to switch between different versions.)

Reviewers:
note that the ruler measurements do not align well with lines in the paper --- this turns out to be very difficult to do well when the paper contains many figures and equations, and, when done, looks ugly.
Just use fractional references (\eg, this line is $087.5$), although in most cases one would expect that the approximate location will be adequate.

\subsection{Paper ID}
Make sure that the Paper ID from the submission system is visible in the version submitted for review (replacing the ``*****'' you see in this document).
If you are using the \LaTeX\ template, \textbf{make sure to update paper ID in the appropriate place in the tex file}.

\subsection{Mathematics}

Please number all of your sections and displayed equations as in these examples:
\begin{equation}
  E = m\cdot c^2
  \label{eq:important}
\end{equation}
and
\begin{equation}
  v = a\cdot t.
  \label{eq:also-important}
\end{equation}
It is important for readers to be able to refer to any particular equation.
Just because you did not refer to it in the text does not mean some future reader might not need to refer to it.
It is cumbersome to have to use circumlocutions like ``the equation second from the top of page 3 column 1''.
(Note that the ruler will not be present in the final copy, so is not an alternative to equation numbers).
All authors will benefit from reading Mermin's description of how to write mathematics:
\url{http://www.pamitc.org/documents/mermin.pdf}.

\subsection{Blind review}

Many authors misunderstand the concept of anonymizing for blind review.
Blind review does not mean that one must remove citations to one's own work---in fact it is often impossible to review a paper unless the previous citations are known and available.

Blind review means that you do not use the words ``my'' or ``our'' when citing previous work.
That is all.
(But see below for tech reports.)

Saying ``this builds on the work of Lucy Smith [1]'' does not say that you are Lucy Smith;
it says that you are building on her work.
If you are Smith and Jones, do not say ``as we show in [7]'', say ``as Smith and Jones show in [7]'' and at the end of the paper, include reference 7 as you would any other cited work.

An example of a bad paper just asking to be rejected:
\begin{quote}
\begin{center}
    An analysis of the frobnicatable foo filter.
\end{center}

   In this paper we present a performance analysis of our previous paper [1], and show it to be inferior to all previously known methods.
   Why the previous paper was accepted without this analysis is beyond me.

   [1] Removed for blind review
\end{quote}

An example of an acceptable paper:
\begin{quote}
\begin{center}
     An analysis of the frobnicatable foo filter.
\end{center}

   In this paper we present a performance analysis of the  paper of Smith \etal [1], and show it to be inferior to all previously known methods.
   Why the previous paper was accepted without this analysis is beyond me.

   [1] Smith, L and Jones, C. ``The frobnicatable foo filter, a fundamental contribution to human knowledge''. Nature 381(12), 1-213.
\end{quote}

If you are making a submission to another conference at the same time, which covers similar or overlapping material, you may need to refer to that submission in order to explain the differences, just as you would if you had previously published related work.
In such cases, include the anonymized parallel submission~\cite{Authors14} as supplemental material and cite it as
\begin{quote}
[1] Authors. ``The frobnicatable foo filter'', F\&G 2014 Submission ID 324, Supplied as supplemental material {\tt fg324.pdf}.
\end{quote}

Finally, you may feel you need to tell the reader that more details can be found elsewhere, and refer them to a technical report.
For conference submissions, the paper must stand on its own, and not {\em require} the reviewer to go to a tech report for further details.
Thus, you may say in the body of the paper ``further details may be found in~\cite{Authors14b}''.
Then submit the tech report as supplemental material.
Again, you may not assume the reviewers will read this material.

Sometimes your paper is about a problem which you tested using a tool that is widely known to be restricted to a single institution.
For example, let's say it's 1969, you have solved a key problem on the Apollo lander, and you believe that the WACV70 audience would like to hear about your
solution.
The work is a development of your celebrated 1968 paper entitled ``Zero-g frobnication: How being the only people in the world with access to the Apollo lander source code makes us a wow at parties'', by Zeus \etal.

You can handle this paper like any other.
Do not write ``We show how to improve our previous work [Anonymous, 1968].
This time we tested the algorithm on a lunar lander [name of lander removed for blind review]''.
That would be silly, and would immediately identify the authors.
Instead write the following:
\begin{quotation}
\noindent
   We describe a system for zero-g frobnication.
   This system is new because it handles the following cases:
   A, B.  Previous systems [Zeus et al. 1968] did not  handle case B properly.
   Ours handles it by including a foo term in the bar integral.

   ...

   The proposed system was integrated with the Apollo lunar lander, and went all the way to the moon, don't you know.
   It displayed the following behaviours, which show how well we solved cases A and B: ...
\end{quotation}
As you can see, the above text follows standard scientific convention, reads better than the first version, and does not explicitly name you as the authors.
A reviewer might think it likely that the new paper was written by Zeus \etal, but cannot make any decision based on that guess.
He or she would have to be sure that no other authors could have been contracted to solve problem B.
\medskip

\noindent
FAQ\medskip\\
{\bf Q:} Are acknowledgements OK?\\
{\bf A:} No.  Leave them for the final copy.\medskip\\
{\bf Q:} How do I cite my results reported in open challenges?
{\bf A:} To conform with the double-blind review policy, you can report results of other challenge participants together with your results in your paper.
For your results, however, you should not identify yourself and should not mention your participation in the challenge.
Instead present your results referring to the method proposed in your paper and draw conclusions based on the experimental comparison to other results.\medskip\\

\begin{figure}[t]
  \centering
  \fbox{\rule{0pt}{2in} \rule{0.9\linewidth}{0pt}}

   \caption{Example of caption.
   It is set in Roman so that mathematics (always set in Roman: $B \sin A = A \sin B$) may be included without an ugly clash.}
   \label{fig:onecol}
\end{figure}

\subsection{Miscellaneous}

\noindent
Compare the following:\\
\begin{tabular}{ll}
 \verb'$conf_a$' &  $conf_a$ \\
 \verb'$\mathit{conf}_a$' & $\mathit{conf}_a$
\end{tabular}\\
See The \TeX book, p165.

The space after \eg, meaning ``for example'', should not be a sentence-ending space.
So \eg is correct, {\em e.g.} is not.
The provided \verb'\eg' macro takes care of this.

When citing a multi-author paper, you may save space by using ``et alia'', shortened to ``\etal'' (not ``{\em et.\ al.}'' as ``{\em et}'' is a complete word).
If you use the \verb'\etal' macro provided, then you need not worry about double periods when used at the end of a sentence as in Alpher \etal.
However, use it only when there are three or more authors.
Thus, the following is correct:
   ``Frobnication has been trendy lately.
   It was introduced by Alpher~\cite{Alpher02}, and subsequently developed by
   Alpher and Fotheringham-Smythe~\cite{Alpher03}, and Alpher \etal~\cite{Alpher04}.''

This is incorrect: ``... subsequently developed by Alpher \etal~\cite{Alpher03} ...'' because reference~\cite{Alpher03} has just two authors.


\begin{figure*}
  \centering
  \begin{subfigure}{0.68\linewidth}
    \fbox{\rule{0pt}{2in} \rule{.9\linewidth}{0pt}}
    \caption{An example of a subfigure.}
    \label{fig:short-a}
  \end{subfigure}
  \hfill
  \begin{subfigure}{0.28\linewidth}
    \fbox{\rule{0pt}{2in} \rule{.9\linewidth}{0pt}}
    \caption{Another example of a subfigure.}
    \label{fig:short-b}
  \end{subfigure}
  \caption{Example of a short caption, which should be centered.}
  \label{fig:short}
\end{figure*}

\section{Method}
\label{sec:formatting}

Prompt learning adapts VL foundational models like CLIP by using learnable prompts instead of full fine-tuning, preserving pre-trained features for specific tasks. However, this approach risks overfitting and reduced generalization to new classes and datasets compared to the original zero-shot CLIP. To address this issue, we propose style-guided Prompt learning (Style-Pro), a new fine-tuning method for vision-language models that reduces the overfitting problem and improves generalization by preventing the trainable model’s embeddings from deviating too far from the pre-trained model’s embedding when learning a new task. Moreover,

\subsection{Preliminaries}
\subsubsection{Contrastive Language-Image Pre-training (CLIP)}
CLIP \cite{radford2021learning} has effectively shown the ability to learn open-set visual concepts. Given an image and its related textual description, CLIP uses visual and text encoders to extract visual and text embeddings. These encoders are pre-trained together with a contrastive objective \cite{oord2018representation} on large-scale image-text pairs to draw related image-text pairs closer and push unrelated pairs apart. Let $f(.)$ and $g(.)$ represent the image encoder and text encoder of the CLIP model, respectively. Their pre-trained parameters are denoted as $\theta_{CLIP} = {\theta_{f},\theta_{g}}$, where $\theta_{f}$ refers to the image encoder parameters and $\theta_{g}$ refers to the text encoder parameters. For a given image, $X$, to obtain the image feature, first, it is divided into $M$ patches followed by a projection called $Patch Embed$ to produce patch tokens. In fact, $Patch Embed$ first splits the input image into fixed-size patches and then projects these patches into features. Then, a learnable class token $CLS$ is concatenated with these features- $\tilde {X} = \{CLS, e_{1}, e_{2}, · · ·, e_{M}\}$, and the concatenated features are sequentially passed through $L$ transformer blocks to produce visual feature representation, $\tilde{\mathbf{f}} \in \mathbb{R}^d$. Similarly, given the ``class name" of $i^{th}$ class, $y$, the $Word Embed$ firstly embeds the hand-crafted description like ''a photo of a \{class name\}" into a vectorized textual tokens which can be formulated as $\tilde {Y} = \{t_{SOS}, t_{1}, t_{2},..., t_{L},c_{k},t_{EOS}\}$. It should be noted that $c_{k}$ represents the word embedding associated with the class label, while $t_{SOS}$ and $t_{EOS}$ denote the learnable embeddings for the start and end tokens, respectively. The text encoder $g$ encodes $\tilde {Y}$ via multiple transformer blocks to produce the latent text feature, $\tilde{\mathbf{g}} \in \mathbb{R}^d$. For zero-shot inference, text features of the text template with class labels $\{1, 2,..., C\}$ are matched with image feature as 
$\frac{\exp(\text{sim}(\tilde{\mathbf{g}}_i, \tilde{\mathbf{f}}) / \tau)}{\sum_{i=1}^C \exp(\text{sim}(\tilde{\mathbf{g}}_i, \tilde{\mathbf{f}}) / \tau)}$, where $\text{sim}()$ denotes the cosine similarity and $\tau$ is the temperature.

\subsubsection{Prompt Learning}
Inspired by prompt learning in NLP, numerous studies have explored the VL models by incorporating learnable prompt tokens during end-to-end training. These learnable prompts can be integrated either on the image side \cite{bahng2022visual}, the text encoder side \cite{zhou2022learning}, or both \cite{khattak2023maple}. In this study, we utilize hierarchical learnable prompt tokens independently for the text and image encoders, following the simple baseline method called Independent Vision-Language Prompting (IVLP) \cite{rasheed2023fine}. 
We concatenate learnable language prompts, denoted as $P_t = \left\{ p_t^1, p_t^2, \cdots, p_t^T \right\}$, and visual prompts, denoted as $P_v = \left\{ p_v^1, p_v^2, \cdots, p_v^V \right\}$, into the respective sets of textual and visual input tokens. Thus, The image encoder generates the prompted visual feature $\tilde{f}_p = f(\tilde{X}_p, \theta_f)$ from input tokens $\tilde{X}p = \{P_v, e{cls}, e_1, e_2, \cdots, e_M\}$, and the textual feature $\tilde{g}_p = g(\tilde{Y}p, \theta_g)$ is obtained from $\tilde{Y}p = \{t_{SOS}, P_t, t_1, t_2, \cdots, t_L, c_k, t_{EOS}\}$. It should be noted that in this work, we employ deep prompting, which involves learning distinct sets of prompts for each transformer block.
\subsection{Type style and fonts}

Wherever Times is specified, Times Roman may also be used.
If neither is available on your word processor, please use the font closest in
appearance to Times to which you have access.

MAIN TITLE.
Center the title $1\frac{3}{8}$ inches (3.49 cm) from the top edge of the first page.
The title should be in Times 14-point, boldface type.
Capitalize the first letter of nouns, pronouns, verbs, adjectives, and adverbs;
do not capitalize articles, coordinate conjunctions, or prepositions (unless the title begins with such a word).
Leave two blank lines after the title.

AUTHOR NAME(s) and AFFILIATION(s) are to be centered beneath the title
and printed in Times 12-point, non-boldface type.
This information is to be followed by two blank lines.

The ABSTRACT and MAIN TEXT are to be in a two-column format.

MAIN TEXT.
Type main text in 10-point Times, single-spaced.
Do NOT use double-spacing.
All paragraphs should be indented 1 pica (approx.~$\frac{1}{6}$ inch or 0.422 cm).
Make sure your text is fully justified---that is, flush left and flush right.
Please do not place any additional blank lines between paragraphs.

Figure and table captions should be 9-point Roman type as in \cref{fig:onecol,fig:short}.
Short captions should be centred.

\noindent Callouts should be 9-point Helvetica, non-boldface type.
Initially capitalize only the first word of section titles and first-, second-, and third-order headings.

FIRST-ORDER HEADINGS.
(For example, {\large \bf 1. Introduction}) should be Times 12-point boldface, initially capitalized, flush left, with one blank line before, and one blank line after.

SECOND-ORDER HEADINGS.
(For example, { \bf 1.1. Database elements}) should be Times 11-point boldface, initially capitalized, flush left, with one blank line before, and one after.
If you require a third-order heading (we discourage it), use 10-point Times, boldface, initially capitalized, flush left, preceded by one blank line, followed by a period and your text on the same line.

\subsection{Footnotes}

Please use footnotes\footnote{This is what a footnote looks like.
It often distracts the reader from the main flow of the argument.} sparingly.
Indeed, try to avoid footnotes altogether and include necessary peripheral observations in the text (within parentheses, if you prefer, as in this sentence).
If you wish to use a footnote, place it at the bottom of the column on the page on which it is referenced.
Use Times 8-point type, single-spaced.

\subsection{Cross-references}

For the benefit of author(s) and readers, please use the
{\small\begin{verbatim}
  \cref{...}
\end{verbatim}}  command for cross-referencing to figures, tables, equations, or sections.
This will automatically insert the appropriate label alongside the cross-reference as in this example:
\begin{quotation}
  To see how our method outperforms previous work, please see \cref{fig:onecol} and \cref{tab:example}.
  It is also possible to refer to multiple targets as once, \eg~to \cref{fig:onecol,fig:short-a}.
  You may also return to \cref{sec:formatting} or look at \cref{eq:also-important}.
\end{quotation}
If you do not wish to abbreviate the label, for example at the beginning of the sentence, you can use the
{\small\begin{verbatim}
  \Cref{...}
\end{verbatim}}
command. Here is an example:
\begin{quotation}
  \Cref{fig:onecol} is also quite important.
\end{quotation}

\subsection{References}

List and number all bibliographical references in 9-point Times, single-spaced, at the end of your paper.
When referenced in the text, enclose the citation number in square brackets, for
example~\cite{Authors14}.
Where appropriate, include page numbers and the name(s) of editors of referenced books.
When you cite multiple papers at once, please make sure that you cite them in numerical order like this \cite{Alpher02,Alpher03,Alpher05,Authors14b,Authors14}.
If you use the template as advised, this will be taken care of automatically.

\begin{table}
  \centering
  {\small{
  \begin{tabular}{@{}lc@{}}
    \toprule
    Method & Frobnability \\
    \midrule
    Theirs & Frumpy \\
    Yours & Frobbly \\
    Ours & Makes one's heart Frob\\
    \bottomrule
  \end{tabular}
  }}
  \caption{Results.   Ours is better.}
  \label{tab:example}
\end{table}

\subsection{Illustrations, graphs, and photographs}

All graphics should be centered.
In \LaTeX, avoid using the \texttt{center} environment for this purpose, as this adds potentially unwanted whitespace.
Instead use
{\small\begin{verbatim}
  \centering
\end{verbatim}}
at the beginning of your figure.
Please ensure that any point you wish to make is resolvable in a printed copy of the paper.
Resize fonts in figures to match the font in the body text, and choose line widths that render effectively in print.
Readers (and reviewers), even of an electronic copy, may choose to print your paper in order to read it.
You cannot insist that they do otherwise, and therefore must not assume that they can zoom in to see tiny details on a graphic.

When placing figures in \LaTeX, it's almost always best to use \verb+\includegraphics+, and to specify the figure width as a multiple of the line width as in the example below
{\small\begin{verbatim}
   \usepackage{graphicx} ...
   \includegraphics[width=0.8\linewidth]
                   {myfile.pdf}
\end{verbatim}
}

\subsection{Color}

Please refer to the author guidelines on the \confName\ \confYear\ web page for a discussion of the use of color in your document.

If you use color in your plots, please keep in mind that a significant subset of reviewers and readers may have a color vision deficiency; red-green blindness is the most frequent kind.
Hence avoid relying only on color as the discriminative feature in plots (such as red \vs green lines), but add a second discriminative feature to ease disambiguation.

\section{Final copy}

You must include your signed IEEE copyright release form when you submit your finished paper.
We MUST have this form before your paper can be published in the proceedings.

Please direct any questions to the production editor in charge of these proceedings at the IEEE Computer Society Press:
\url{https://www.computer.org/about/contact}.

{\small
\bibliographystyle{ieee_fullname}
\bibliography{egbib}
}